\documentclass{llncs}

\usepackage[pdftex]{graphicx}
\usepackage{amsmath}
\usepackage{amssymb}
\usepackage{amsfonts}
\usepackage{floatrow}

\usepackage{multirow}
\usepackage{rotating}

\usepackage{subfigure}
\usepackage{color}
\usepackage{ogonek}

\definecolor{myred}{rgb}{.8,.0,.0}

\begin{document}

\title{Label Stability in Multiple Instance Learning}


\author{Veronika Cheplygina$^{1,3}$, Lauge S{\o}rensen$^2$, David M. J. Tax$^1$,\\ Marleen de Bruijne$^{2,3}$ and Marco Loog$^{1,2}$}


\institute{$^1$ Pattern Recognition Laboratory, Delft University of Technology, The Netherlands \\
$^2$ The Image Section, University of Copenhagen, Copenhagen, Denmark \\
$^3$ Biomedical Imaging Group Rotterdam, Erasmus MC, Rotterdam, The Netherlands
}

\maketitle

\begin{abstract}
We address the problem of \emph{instance label stability} in multiple instance learning (MIL) classifiers. These classifiers are trained only on globally annotated images (bags), but often can provide fine-grained annotations for image pixels or patches (instances). This is interesting for computer aided diagnosis (CAD) and other medical image analysis tasks for which only a coarse labeling is provided. Unfortunately, the instance labels may be unstable. This means that a slight change in training data could potentially lead to abnormalities being detected in different parts of the image, which is undesirable from a CAD point of view.  Despite MIL gaining popularity in the CAD literature, this issue has not yet been addressed. We investigate the stability of instance labels provided by several MIL classifiers on 5 different datasets, of which 3 are medical image datasets (breast histopathology, diabetic retinopathy and computed tomography lung images). We propose an unsupervised measure to evaluate instance stability, and demonstrate that a performance-stability trade-off can be made when comparing MIL classifiers.
\end{abstract}

\section{Introduction}

Obtaining ground-truth annotations for patches, which can be used to train supervised classifiers for localization of abnormalities in medical images can be very costly and time-consuming. This hinders the use of supervised classifiers for this task. Fortunately, global labels for whole images, such as the overall condition of the patient, are available more readily. Multiple instance learning (MIL) is an extension of supervised learning which can train classifiers using such weakly labeled data. For example, a classifier trained on images (\emph{bags}), where each bag is labeled as healthy or abnormal and consists of unlabeled image patches (\emph{instances}), would be able to label patches of a novel image as healthy or abnormal.

MIL is becoming more and more popular in CAD~\cite{kandemir2014computer,melendez2014novel,cheplygina2014classification,wang2012seeing,quellec2012multiple,bi2007multiple,wu2009min,sun2012ecg,marques2013osteoarthritis}. In many of these applications, it is desirable to obtain instance labels, and to inspect the instances which are deemed positive. For example, in~\cite{melendez2014novel}, weakly labeled x-ray images of healthy subjects and patients affected by tuberculosis are used to train a MIL classifier which can provide local abnormality scores, which can be visualized across the lungs. Furthermore, the MIL classifier \emph{outperforms its supervised counterpart} which has access to fine-grained labels, showing the potential of MIL for CAD applications.

A pitfall in using MIL classifiers to obtain instance labels is that these labels might be unstable, for example, if a different subset of the data is used for training. This is clearly undesirable in a diagnostic setting, because abnormalities would be highlighted in different parts of the image. For example, in \cite{marques2013osteoarthritis} a MIL classifier is used to identify which of the 8 regions (instances) of the tibial trabecular bone (bag) are most related to cartilage loss. The ``most positive'' region labeled positive by only 20\% of the classifiers, trained on different subsets of the data.  We have not been able to identify other research where this phenomenon is investigated, which emphasizes the importance of the present work.

In rare cases where instance-level annotations are available, such as in~\cite{kandemir2014computer}, instance labels can be evaluated using AUC. The results here show that the best bag classifier does not correspond to the best instance classifier, emphasizing that bag-level results are not reliable if instance labels are needed. Another approach is to evaluate the instances qualitatively. However, this is typically done for a single run of the classifier, which raises the question whether the same abnormalities would be found if the training set would change slightly.

We propose to evaluate the \emph{stability} of instance-labeling MIL classifiers as an additional measure for classifier comparison. We evaluate two stability measures on three CAD datasets: computed tomography lung images with chronic obstructive pulmonary disease (COPD), histopathology images with breast cancer and diabetic retinopathy images.  We demonstrate how stability varies in popular MIL classifiers, and show that choosing the classifier with the best bag-level performance may not lead to reliable instance labels.


\section{Multiple Instance Learning}\label{sec:mil}


In multiple instance learning, a sample is a bag or set $B_i  = \{\mathbf{x}^i_{k}| k=1,...,n_i\} \subset \mathbb{R}^d$ of $n_i$ instances, each instance is thus a $d$-dimensional feature vector. We are given labeled training bags $\{(B_i, y_i) | i=1,...N_{tr}\}$ where $y_i \in \{0, 1\}$. The standard assumption is that there exist hidden instance labels $z^i_{k} \in \{0, 1\}$ which relate to the bag labels as follows: a bag is positive if and only if it contains at least one positive instance.

Originally, the goal in MIL is to train a bag classifier $f_B$ to label previously unseen bags. Several MIL classifiers do this by inferring an instance classifier $f_I$, and combining the outputs of the bag's instances, for example by the noisy-or rule, $f_B(B_i) = \max_k\{f_I(\mathbf{x}^i_k)\}$. An example of such an \emph{instance-level} classifier is SimpleMIL, which propagates the bag label to its instances and simply trains a supervised classifier on the, possibly noisy, instance labels. Classifiers which explicitly use the MIL assumption are miSVM~\cite{andrews2002support} and milBoost~\cite{viola2006multiple}, which are MIL adaptations of popular learning algorithms. For example, miSVM extends the SVM by not only searching for the optimal hyperplane $\mathbf{w}$ which defines $f_I$, but also for the instance labels $\{z_i^k\}$ which are consistent with the bag label assumptions:

\begin{eqnarray}
\min_{\{z_i^k\}} \min_{\mathbf{w},\xi} \frac{1}{2} ||\mathbf{w}||^2 + C \sum_{i,k} \xi_i^k & \text{s.t.} \\
\forall i,k: z_i^k(\langle \mathbf{w},\mathbf{x}_i^k \rangle) \geq 1 - \xi_i^k, & \xi_i^k \geq 0, z_i^k \in \{-1,1\}, \max{\{z_i^k\}} = y_i. \nonumber
\end{eqnarray}

Another group, \emph{bag-level} classifiers, typically represent each bag as a single feature vector and use supervised classifiers for training $f_B$ directly~\cite{chen2006miles,cheplygina2014multiple}. Such classifiers are often robust, but usually can not provide instance labels. A notable exception is MILES~\cite{chen2006miles}, which represents each bag by its similarities to a set of prototype instances, $\mathbf{s}_i = [s(B_i,\mathbf{x}^1_1), \ldots, s(B_i, \mathbf{x}^1_{n_1}), \ldots s(B_i, \mathbf{x}^{N_{tr}}_{n_{N_{tr}}})]$ where $s(B_i,\mathbf{x}) = \exp(- \min_{k} ||\mathbf{x} - \mathbf{x}^i_{k} ||)$ or any other kernel. A sparse classifier then selects the most discriminative features, which correspond to instance prototypes. It is assumed that discriminative prototypes from positive bags are positive, instances can therefore be classified based on their similarity to these prototypes. 

The interest in \textbf{MIL for computer aided diagnosis} has grown over the past decade, as illustrated by Table~\ref{tab:applications}. Supervised evaluation of instances is only performed in a few studies -- where (a part) of the data has been annotated at the instance level. Otherwise, papers examine the instances qualitatively, such as displaying the most abnormal instances~\cite{melendez2014novel}, or not at all, although instance labels would be interesting from a diagnostic point of view~\cite{cheplygina2014classification,kandemir2014computer}. As our proposed evaluation is unsupervised, it can easily be adopted in all these studies.

\begin{table}
\begin{tabular}{l l l l l l }

Task & B & I & Task & B & I \\

\hline

Cancer histology~\cite{kandemir2014computer}   & + & + & Diabetic retinopathy~\cite{kandemir2014computer} & +  & -- \\

COPD in CT~\cite{cheplygina2014classification} & + & -- & Tuberculosis in XR~\cite{melendez2014novel}  &  + & $\circ$ \\

Cancer histopathology~\cite{xu2014weakly}   &  + & + & Osteoarthritis in MRI ~\cite{marques2013osteoarthritis}  & +  & $\circ$  \\

Diabetic retinopathy~\cite{quellec2012multiple}  &   +  &  +  &  Pulmonary embolism in CT ~\cite{liang2007computer}	& + & -- \\

Myocardial infarction in ECG~\cite{sun2012ecg}  &    +  &  -- & COPD in CT~\cite{sorensen2012texture}   & +  & --					  \\    

Colorectal cancer in CT~\cite{dundar2008multiple}    &  + & --  & & & \\



\hline

\end{tabular}
\caption{Evaluation of MIL in CAD tasks. Columns show bag (B) and instance (I) evaluation: supervised (+), qualitative ($\circ$) or none (--).}
\label{tab:applications}
\end{table}

\section{Instance Stability}

We are interested in evaluating the similarity of a labeling, or vector of outputs of two classifiers $\mathbf{z} = f_I(X)$ and $\mathbf{z}^\prime = f_I^\prime(X)$, trained on slightly different subsets of the training data, for the test set $X = [\mathbf{x}^1_1, \ldots , \mathbf{x}^N_{n_N}]^\intercal$. The stability measure should be \textbf{monotonically increasing} with the number of instances the classifiers agree on, have \textbf{limits}, and most importantly, be \textbf{unsupervised}, i.e. not dependent on the hidden instance labels $z_i$.

The general concept of stability is important in machine learning, and different aspects of it have been addressed in the literature. Leave-one-out stability~\cite{poggio2004general} measures to what extent a decision boundary changes when a sample is removed from the training data, but is not appropriate because it is supervised. An unsupervised version where bags are left out, and true labels are substituted by classifier outputs, is related to the measures we propose. The kappa statistic is unsupervised, but does not follow the monotonicity property in class imbalance settings which could occur in MIL. Clustering stability~\cite{ben2001stability} compares the outputs of two clustering procedures and is unsupervised. It is appropriate for our goal and is in fact related to the measures proposed in what follows.

Let $n_{00} = |\{i | z_i = 0 \land z^\prime_i = 0 \}|$,  $n_{01} = |\{i | z_i = 0 \land z^\prime_i = 1 \}|$  ,  $n_{10} = |\{i | z_i = 1 \land z^\prime_i = 0 \}|$ and  $n_{11} = |\{i | z_i = 1 \land z^\prime_i = 1 \}|$. An intuitive measure that satisfies the properties above is the agreement fraction:

\begin{equation}
S(\mathbf{z}, \mathbf{z}^\prime) = (n_{00} + n_{11}) / (n_{01} + n_{10} + n_{11} + n_{00}).
\end{equation}

In a situation with many true negative instances, the value of $S$ would be inflated due to the negative instances that the classifiers agree on. As a result, the classifier can still be unstable with respect to the positive instances. Due to the nature of CAD tasks, we might consider it more important for the classifiers to agree on the positive instances. Therefore we also consider the agreement on positive labels only, or Jaccard distance:

\begin{equation}
S_+(\mathbf{z}, \mathbf{z}^\prime) =  n_{11} / ( n_{01} + n_{10} + n_{11} ).
\end{equation}

We emphasize that the novelty does not lie in the measures themselves, well-known as they are. The novelty resides in what they measure in this context: we derive these measures as the appropriate ones for the stability that we want to quantify.

\textbf{Classifier Selection.} If instance classification stability is a crucial issue, one can study our measure in combination with bag-level AUC (or any other accuracy measure) and select a MIL classifier with a good trade-off of AUC and instance stability. We can see each classifier as a possible solution, parametrized by these two values. Intermediate solutions between classifiers $f_I$ and $f_I^\prime$ can in theory be obtained by designing a randomized classifier, which trains classifier $f_I$ with probability $p$ and classifier $f_I^\prime$ with probability $1-p$. In the AUC-stability plane, the Pareto frontier is the set of classifiers which are Pareto efficient, i.e. no improvement can be made in AUC without decreasing instance stability and vice versa. Optimal classifiers can therefore be selected from this Pareto frontier. While the classifier with the highest AUC is in this set, it is not necessarily the only desirable solution, if the instance labels are of importance.

\section{Experiments and Results}~\label{sec:experiments}
\textbf{Datasets.} The datasets are shown in Table~\ref{tab:data}. The Musk datasets are benchmark problems of molecule activity prediction. In Breast, an instance is a $7\times7$  patch from a $896 \times 768$ tissue microarray analysis image from a patient with a malignant (+) or benign (--) tumor. In Messidor, an instance is a $135 \times 135$ patch from a $700 \times 700$ fundus image of a diabetes (+) or healthy (--) subject. In COPD, a bag is a CT image of a lung of a subject with COPD (+) or a healthy subject (--). An instance is a region of interest (ROI) of $41 \times 41\times 41$ voxels, with the center inside the segmentation of the lung field. 

\begin{table}%
\centering
\begin{tabular}{l l p{1.5cm} p{2cm}  l l  }

Dataset & Bags & Instances & Inst per bag & Features \\
\hline

Musk 1 &  47+, 45-- & 476 & 2 to 40 &  166  \\
Musk 2 &  39+, 63-- & 6598 & 1 to 1024 &   166  \\
Breast~\cite{kandemir2014empowering}  &  26+, 32--  &  2002 & 21 to 40   & 657 (intensity, LBP, SIFT) \\
Messidor~\cite{kandemir2014computer,decenciere2014feedback} &  654+, 546--  & 12352 & 8 to 12 & 687 (intensity, LBP, SIFT)\\
COPD~\cite{sorensen2012texture,pedersen2009danish}       &   231+, 231--  & 26200    & 50 &  287 (Gaussian filter bank) \\
                                          \hline

\end{tabular}
\caption{Datasets and their properties. Musk, Breast and Messidor can be downloaded from a MIL data repository~\cite{cheplygina2014multiple} (\url{http://www.miproblems.org}).}
\label{tab:data}
\end{table}

\begin{figure}
\floatbox[{\capbeside\thisfloatsetup{capbesideposition={right},capbesidewidth=4cm}}]{figure}[\FBwidth]
{\caption{Pairwise stability for 10 MILES classifiers for agreement (left) and positive agreement (right) for the COPD dataset.}\label{fig:toy}}
{\includegraphics[width=0.55\textwidth]{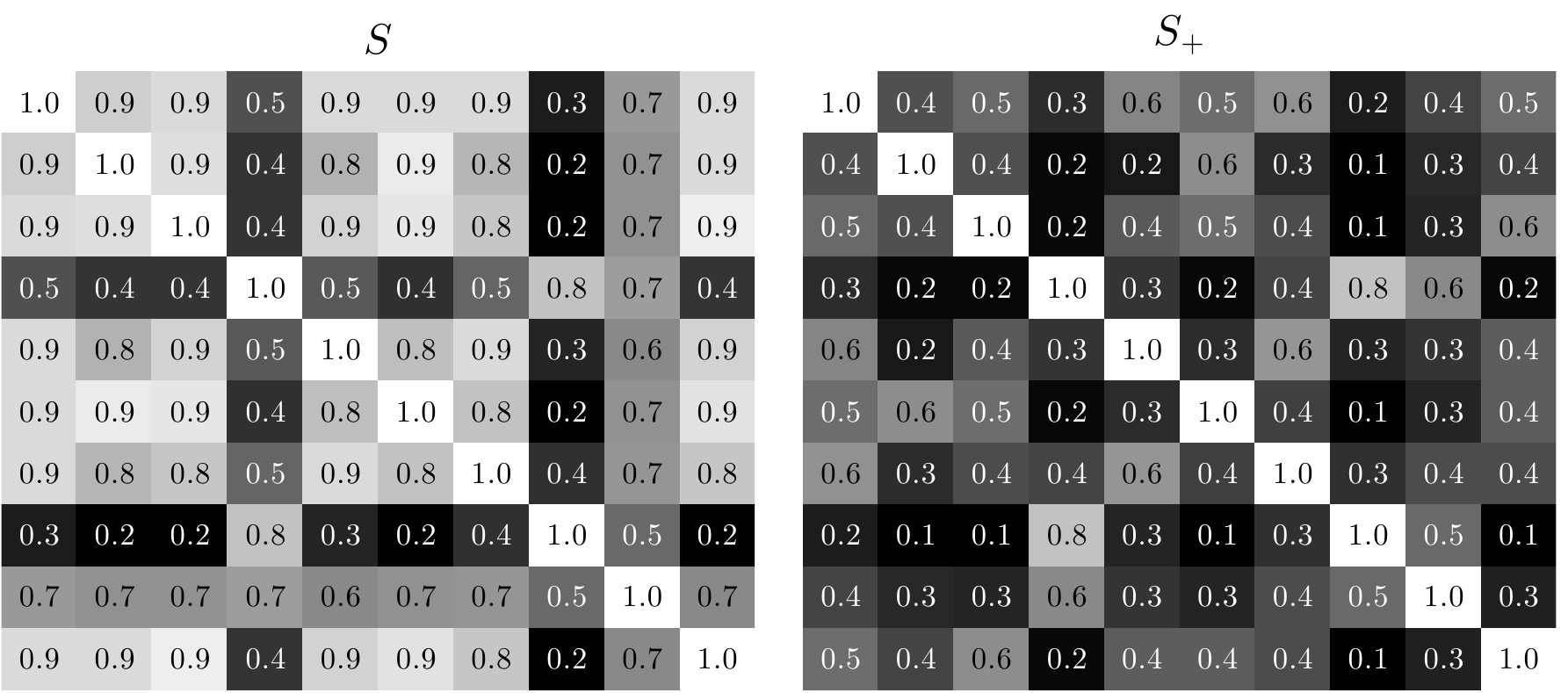}}
\end{figure}

\textbf{Illustrative Example.} Fig.~\ref{fig:toy} shows the pairwise stability measures for the COPD validation data, for 10 MILES classifiers, each trained on random 80\% of the training data. There is considerable disagreement for both measures, which is surprising because of the large overlap of the training sets. The measures are quite correlated ($\rho = 0.76$), but $S$ has higher values because it is inflated by agreement on negative instances. 


\begin{figure}%
\centering
\includegraphics[width=0.4\textwidth]{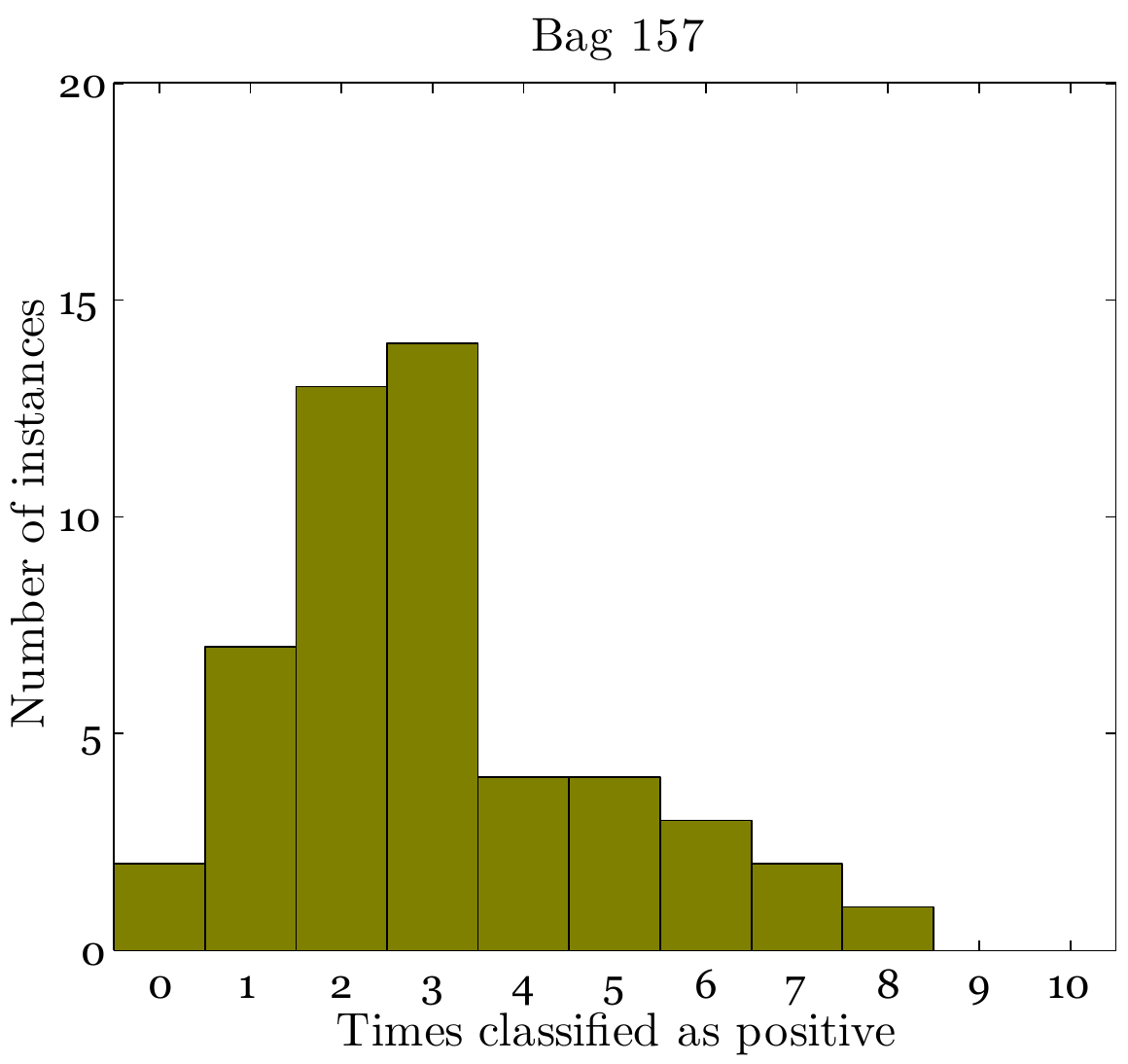}%
\raisebox{5mm}{\includegraphics[width=0.50\textwidth]{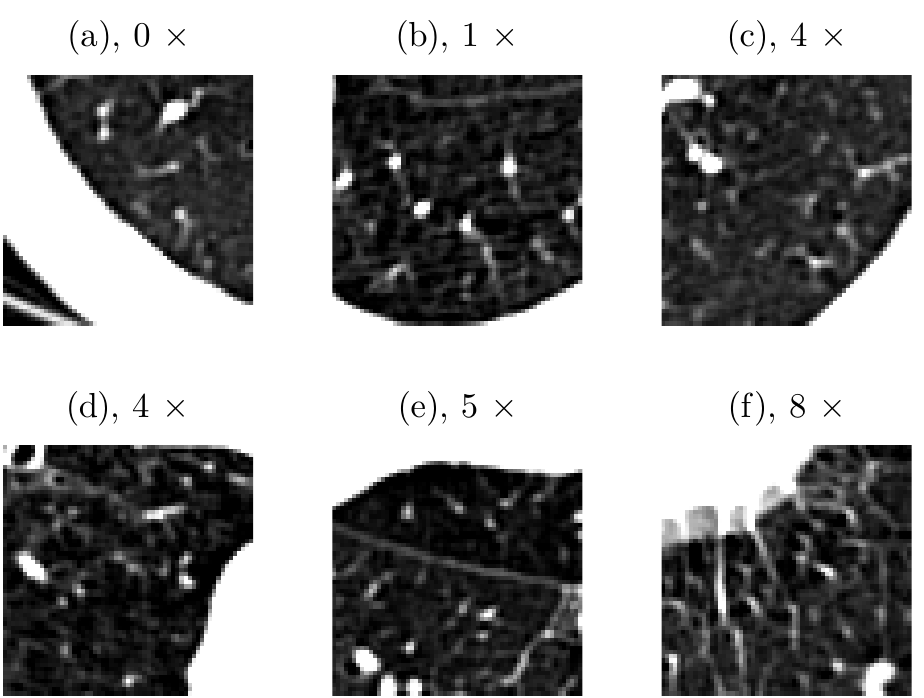}}%

\caption{``Positiveness'' of 50 instances (ROIs) from a positive bag. \textbf{Left:} How often an ROI is classified positive, and for how many ROIs this holds. \textbf{Right}: Examples of 6 ROIs, for which the axial slice with most lung voxels below -910 hounsfield units is shown. ROIs (b,d,e,f) contain emphysema (low intensity areas within the lung tissue), but only (f) is often classified as positive. ROIs (a,c) are largely unaffected, but only (a) is consistently classified as negative.}
\label{fig:inst}%
\end{figure}

Fig.~\ref{fig:inst} shows how the instance classifications change in a true positive bag, i.e., CT image from a COPD patient. This bag is always classified as positive, but the instance labels are unstable. A perfectly stable classifier would have a bimodal distribution, classifying instances as positive either 0 or 10 times. We also show a number of ROIs with stable and unstable labels. Several ROIs containing emphysema have unstable classifications, and one emphysemous patch is even consistently classified as negative. This shows that while the bag is always classified correctly, the instance labels may not be very reliable.

\textbf{Evaluation.} We evaluate a number of classifiers (please see Sec.~\ref{sec:mil} for descriptions) from the MIL toolbox\footnote{http://prlab.tudelft.nl/david-tax/mil.html}, which we modified to output instance labels:

\begin{itemize}
\item simpleMIL with SVM, nearest mean (NM) and 1-nearest neighbor (1NN)
\item miSVM and its variants miNM and mi1NN (based on NM and 1NN)
\item MILBoost
\item MILES
\end{itemize}
We use a linear kernel and regularization parameter $C=1$ for SVM, miSVM and MILES. For each train/test split, we do the following 10 times: randomly sample 80\% of the training bags (bag = subject), train the classifier, and evaluate on the test dataset. The splits are done randomly for Musk, Breast and Messidor and based on predefined sets for COPD.

\begin{figure}%
\centering

\includegraphics[width=0.80\textwidth]{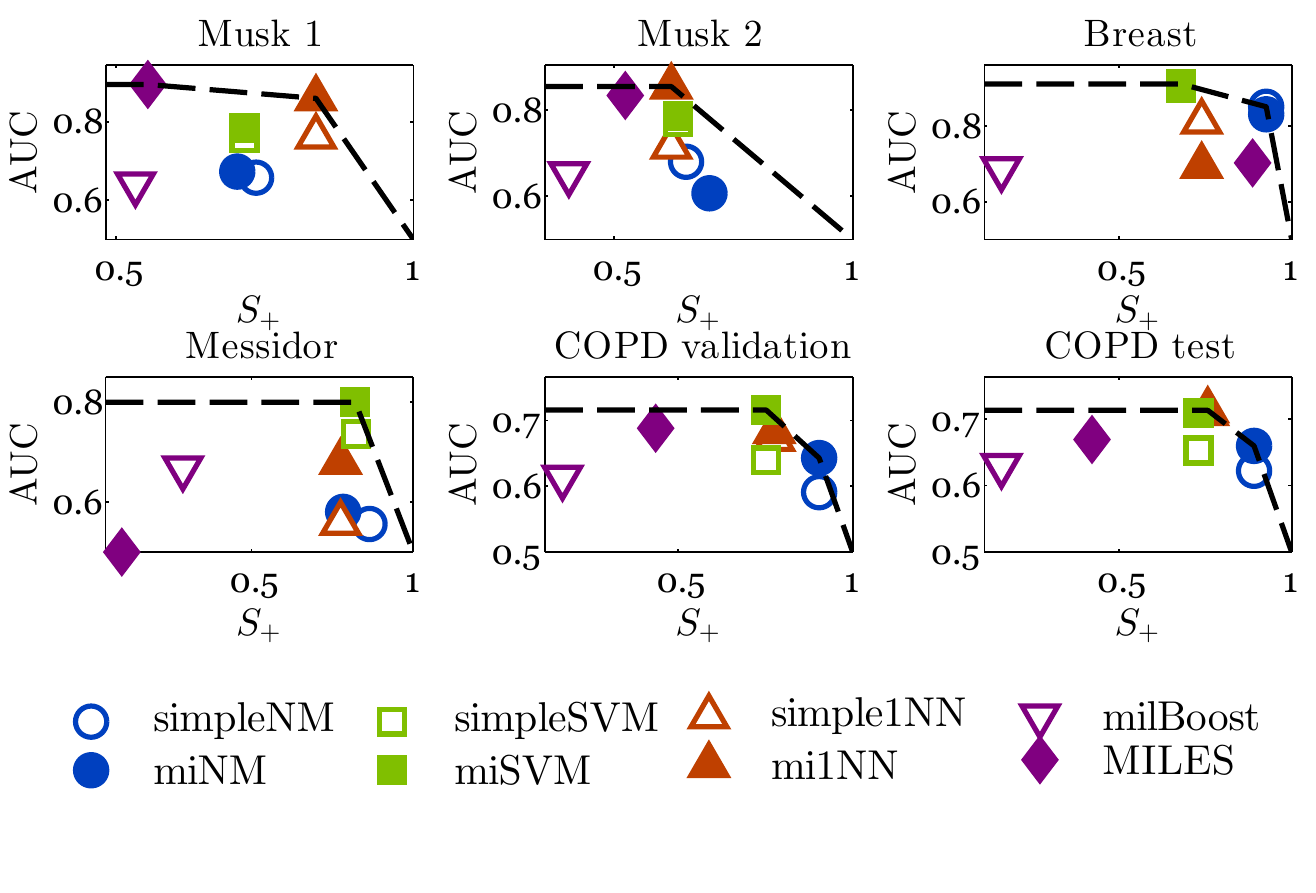}%

\caption{Bag AUC vs positive instance stability and the corresponding Pareto frontiers.}%
\label{fig:stability1}%
\end{figure}

The average bag AUCs, average pairwise instance stabilities, and the corresponding Pareto frontiers for $S+$ ($S$ provided similar plots, but with inflated values) are shown in Fig.~\ref{fig:stability1}.  Note that the (1, 0.5) point can be achieved by a classifier which labels all instances as positive. The main observation is that the most accurate classifier is often not the most stable one. This trade-off is especially well-illustrated in the COPD datasets. Here we see similar behavior between the two sets, which shows that if we were to use the validation set results for classifier selection, we would obtain a classifier with similar performance and stability on the test set.

With regard to the classifiers, miSVM and its variants seem to be relatively good choices. MILES, which is a popular classifier due to its good performance, can indeed be quite accurate, but at the same time unstable. The difference between the mi- classifiers and MILES is probably due to the fact that MILES trains a bag classifier $f_B$ first, and infers $f_I$ from $f_B$, while the mi- classifiers train $f_I$ directly. MILBoost is both inaccurate and unstable, especially for COPD there is high disagreement on which instances to label as positive.

Note that the goal of these experiments is to demonstrate the trade-off between AUC and stability, not to maximize the AUC. Nevertheless, the best performances achieved by classifiers tested here are [0.91, 0.80, 0.72, 0.72] for Breast, Messidor, and the COPD datasets. In previous works, the highest\footnote{Note that \cite{cheplygina2014classification} reports several higher AUCs for COPD, but these correspond to a larger version of the dataset, which was not used in our study} performances for the same datasets were [0.90, 0.81, 0.74, 0.74]. This shows that our result are on par with state of the art, despite using less data and optimization.


\section{Conclusions}

We addressed the issue of stability of instance labels provided by MIL classifiers. We examined two unsupervised measures of agreement: $S$ based on all labels, and $S_+$ based on positive (abnormal) labels, which might be more interesting from a CAD point of view. Our experiments demonstrate a trade-off between bag performance and instance label stability, and miSVM is a classifier which provides a good trade-off. In general, we propose to use instance label stability as an additional evaluation measure when applying MIL classifiers in CAD.


\textbf{Acknowledgements}. This research is partially financed by The Netherlands Organization for Scientific Research (NWO). We thank Dr. Melih Kandemir for kindly providing the Breast and Messidor datasets.

\bibliographystyle{splncs03}
\bibliography{refs}

\end{document}